\documentclass[11pt,a4paper]{article}
\usepackage[hyperref]{acl2021}
\usepackage{times}
\usepackage{latexsym}
\usepackage{booktabs}
\usepackage{ threeparttable }
\usepackage{multirow}
\usepackage{graphicx}
\usepackage{algorithm}
\usepackage{algpseudocode}
\usepackage{algpseudocode}

\newcommand*{\thead}[1]{\multicolumn{1}{c}{\bfseries #1}}

\usepackage{microtype}

\aclfinalcopy 

\title{BME Submission for SIGMORPHON 2021 Shared Task 0. A Three Step Training Approach with Data Augmentation for Morphological Inflection}
\author{Gábor Szolnok$^*$ \\
  Budapest University of  \\
  Technology and Economics \\
  \texttt{szolnokgabcsi@gmail.com} \\\And
  Botond Barta$^*$ \\
  Budapest University of  \\
  Technology and Economics \\
  \texttt{b.botond25@gmail.com} \\\AND
  Dorina Lakatos$^*$ \\
  Budapest University of  \\
  Technology and Economics \\
  \texttt{dorinapetra@gmail.com} \\\And
  Judit Ács \\
  Budapest University of  \\
  Technology and Economics \\
  \texttt{judit@sch.bme.hu} \\}
  
\date{}

\begin{document}
\maketitle
\begin{abstract}

We present the BME submission for the SIGMORPHON 2021 Task 0 Part 1, Generalization Across Typologically Diverse Languages shared task.
We use an LSTM encoder-decoder model with three step training that is first trained on all languages, then fine-tuned on each language families and finally fine-tuned on individual languages.
We use a different type of data augmentation technique in the first two steps.
Our system outperformed the only other submission.
Although it remains worse than the Transformer baseline released by the organizers, our model is simpler and our data augmentation techniques are easily applicable to new languages.
We perform ablation studies and show that the augmentation techniques and the three training steps often help but sometimes have a negative effect.
Our code is publicly available\footnote{\url{https://github.com/szogabaha/SIGMORPHON2021-task0}\\$^*$ The first three authors contributed equally.}.

\end{abstract}

\section{Introduction}

Morphological inflection is the task of generating inflected word forms given a lemma and a set of morphosyntactic tags.
Inflection plays a key role in natural language generation, particularly in languages with rich morphology.
The SIGMORPHON Shared Tasks are yearly competitions for inflection tasks\cite{cotterell2016sigmorphon, cotterell-conll-sigmorphon2017, cotterell-etal-2018-conll, mccarthy-etal-2019-sigmorphon, sigmorphon-2020-sigmorphon}.

This paper describes the BME team's submission for Part 1 of the 2021 SIGMORPHON--UniMorph Shared Task on Generalization in Morphological Inflection Generation.
There were only two submissions to this subtask and our team outperformed the other submission by a large margin.
The task was about type-level morphological inflection in 38 typologically diverse languages from 12 language families.

Our model builds on the work of \citet{Faruqui:2015}.
We use a sequence-to-sequence (seq2seq) model with a bidirectional LSTM \cite{Hochreiter:1997} encoder and a unidirectional LSTM decoder with attention.
We perform a small hyperparameter search and we find that the most important parameters are the choice of the embedding size and the hidden size.

We use two data augmentation techniques; a simple copy mechanism, a stem modification method.
We add these methods at the first two steps of our three-step training regime.
We describe a third data augmentation technique, a template induction method, that did not improve the overall results in the end so we did not use it in our submission. 
We first train a single model on all languages and then fine-tune the model on each language family and then on each language.

Our main contributions are:
\begin{itemize}
    \item We present the highest performing submission for the SIGMORPHON 2021 Task0 Part 1 shared task.
    \item We try three data augmentation techniques.
    \item We use a three-step training regime mixed with a two data augmentation techniques applied at the first two steps.
    \item We perform ablation studies for the augmentation techniques and the training steps. We highlight the negative results as well.
\end{itemize}

\section{Related Work}

Seq2seq neural networks were first popularized in machine translation \cite{Sutskever_2014} and since the addition of the attention mechanism \cite{bahdanau2014neural, luong-etal-2015-effective} and the Transformer \cite{vaswani}, it has been the dominant approach in the field.
The first major success of seq2seq models in morphological inflection was the submission by \citet{Kann:2016} to the 2016 edition of the SIGMORPHON shared task.
This was followed by an extensive study by \citet{Faruqui:2015} on LSTM-based encoder-decoder models for morphological inflection.

We used the augmentation techniques introduced by \citet{Neuvel:2002}. Inspired by \citet{bergmanis-etal-2017-training} we attempted to extract different morphological properties of the languages and used them to generate data.
\citet{anastasopoulos-neubig-2019-pushing} used a two step training method that first trains on the language family and then on the individual languages.
We use a similar procedure but we augment the data with a different technique in each step.

\section{Data}

The shared task covered 38 languages from 12 language families.
35 of these languages were available from the beginning while 3 surprise languages, Turkish, Vibe and Võro, were released one week before the submission deadline.
Each language had a train and a development split of varying size.
Each sample consists of a lemma, an inflected form and a list of morphosyntactic tags in the following format:\\
\texttt{\small{
vaguear	vaguearás	V;IND;SG;2;FUT \\
emunger	emunjamos	V;IMP;PL;1;POS \\
desenchufar	desenchufo	V;IND;SG;1;PRS \\
delirar	deliraren	V;SBJV;PL;3;FUT}}

The amount of data varies widely among the languages: while the language Veps has more than 100000 examples, Ludic, the most underresourced language, has only 128 train samples.
Table~\ref{tab:family} lists the 12 language families and the number of languages from that family.
We consider some languages low resource languages if they have fewer than 1300 samples.
8 language families had at least one low resource language and 3 families were represented only by low resource languages.
One goal of our data augmentation techniques is to offset this imbalance (see Section~\ref{Augmentation}).

\begin{table}[h]
\centering
\resizebox{0.49\textwidth}{!}{
\begin{tabular}{lrrr}
\toprule
 \thead{\textbf{Family}} & \thead{\textbf{Langs}} & \thead{\textbf{Low}} & \thead{\textbf{Samples}} \\
 \midrule
 Afro-Asiatic & 12 & 1 & 196550 \\
 Arnhem & 1 & 1 & 214 \\
 Aymaran & 1 & 0 & 100000 \\
 Arawakan & 2 & 0 & 16472 \\
 Iroquoian & 1 & 0 & 3801 \\
 Turkic & 3 & 0 & 300371 \\
 Chukotko-Kamchatkan & 2 & 2 & 1378 \\
 Tungusic & 2 & 1 & 5500 \\
 Austronesian & 2 & 1 & 11395 \\
 Trans-New-Guinean  & 1 & 1 & 918 \\
 Indo-European & 12 & 2 & 685567 \\
 Uralic & 5 & 2 & 279720 \\
 \bottomrule
\end{tabular}
}
\caption{List of language families and the number of languages from each family. The third column is the number of low resource (\textless 1300 samples) languages in a particular family. The forth column is the overall sample count in each family.} 
\label{tab:family}
\end{table}

\section{Model architecture}\label{sec:model}

\subsection{LSTM based seq2seq model}

Our model is largely based on the encoder-decoder model of \citet{Faruqui:2015}.
We use a bidirectional LSTM \cite{Hochreiter:1997} as our encoder and a unidirectional LSTM with attention as our decoder.

Recall that the input for the inflection task is a pair consisting of a lemma and a list of morphosyntactic tags.
We represent these pairs as a single sequence as the LSTM's input.
For the input lemma-tags pair \texttt{izar, (V, COND, PL, 2)}, we serialize it as

{\centering \texttt{\small{<SOS> i z a r <SEP> V COND PL 2 <EOS>}}\par }

Similarly, we convert the target form into a sequence of characters:

{\centering \texttt{\small{<SOS> i z a r <EOS>}}\par }

The output of our model looks like this when the inflected word is izaríais:

{\centering\texttt{\small{<SOS> i z a r í a i s <EOS>}} \par}

The input sequence is first projected to an embedding space, which then provides the input for the encoder LSTM.
The decoder is a standard unidirectional LSTM with attention.
We decode the output in a greedy fashion and do not use beam search.
We project the final output to the output vocabulary's dimension and use the softmax function to generate a probability distribution.
The input and the output embeddings use shared weights and they are trained from scratch along with the rest of the model.

%
%
%
%


\begin{table*}[t]
\begin{threeparttable}
\centering
\begin{tabular}{ll|l|rrrrr|rr}
\multicolumn{1}{c}{\multirow{2}{*}{\textbf{Family}}}                      
& \multirow{2}{*}{\begin{tabular}[c]{@{}l@{}}\textbf{Lang }\\\textbf{code}\end{tabular}} 
& \multirow{2}{*}{\textbf{Result}} & \multicolumn{5}{c|}{\textbf{excluded feature}}                                                     & \multicolumn{2}{c}{\textbf{basemodels}}              \\
\multicolumn{1}{c}{}  &   &                                  
& \multicolumn{1}{c}{\textit{Copy}}                                                                                            &\multicolumn{1}{c}{\textit{Stem-mod}} & \multicolumn{1}{c}{\textit{Step 1}} & \multicolumn{1}{c}{\textit{Step 2}} & \multicolumn{1}{c|}{\textit{Step 3}} & \multicolumn{1}{c}{IIT+DA} & \multicolumn{1}{c}{OL}  \\ 
\hline
Turkic                  & tur   & 99.90 & 99.90 & 99.94 & 99.92 & 97.38 & 99.90 & 99.35 & 97.10 \\ \hline
\multirow{3}{*}{Uralic} & vep   & 99.72 & 54.10 & 99.55 & 99.80 & 99.05 & 99.67 & 99.70 & 91.13 \\
                        & lud   & 59.46 & 56.76 & 70.27 & 56.76 & 67.57 & 62.16 & 45.95 & 0.00  \\
                        & olo   & 99.72 & 91.15 & 99.84 & 99.78 & 98.26 & 99.72 & 99.66 & 99.48 \\ \hline
\multirow{3}{*}{\begin{tabular}[c]{@{}l@{}}Indo -\\European\end{tabular}} 
                        & rus   & 98.07 & 94.84 & 98.00 & 97.86 & 95.56 & 97.34 & 97.58 & 70.72  \\
                        & kmr   & 98.21 & 86.02 & 98.74 & 98.41 & 97.50 & 98.21 & 98.01 & 5.14   \\
                        & deu   & 97.98 & 91.19 & 98.23 & 97.91 & 89.91 & 97.98 & 97.46 & 91.86                  
\end{tabular}
\caption{The different results we achieved on the test dataset with different models, with different augmentation techniques excluded and with different training steps excluded. For comparison the table show the result of our submission (\emph{result}), the given basemodel \emph{IIT+DA} (Input Invariant Transformer + Data Augment) and the models that were just trained on only one language (\emph{OL}). }
\label{tab:results}
\end{threeparttable}
\end{table*}

\subsection{Hyperparameter selection}

We selected 16 languages from diverse families for hyperparameter tuning.
Most of them were fusional or low resource because early experiments showed that these are the harder ones to learn for the model.
We downsampled the larger languages and merged the train sets.
We trained 100 models with random parameters sampled uniformly from the parameter ranges listed in Table~\ref{tab:param_ranges}.

\begin{table}[ht]
    \centering
    \begin{tabular}{crrr}
    \toprule
    \textbf{Parameter} & \textbf{Type} & \textbf{Min value} & \textbf{Max value} \\
    \midrule
         num.~layers & int & 1 & 3 \\
         dropout & float & 0.1 & 0.6 \\
         embedding & int & 32 & 256 \\
         hidden size & int & 64 & 1024 \\
     \bottomrule
    \end{tabular}
    \caption{Parameter ranges used for hyperparameter tuning.}
    \label{tab:param_ranges}
\end{table}

It turns out that only two of these hyperparameters makes a significant difference, the number of layers and the hidden size.
One LSTM layer was clearly inferior and so were LSTMs with fewer than 400 neurons per layer.
The embedding dimension and the dropout rate made less difference.
We decided to go with two configurations, a small one with 200 dimensional embedding and the hidden size set to 256 and a large one with the embedding set to 150 dimensions and 900 hidden size.
We report the better one for each language based on the development set.

\subsection{Training details}

We train the models end-to-end with gradient descent using the Adam optimizer with 0.001 learning rate.
We apply teacher forcing to the decoder with 0.5 probability, which means that we feed the ground truth character instead of the output of the previous step half of the time.

\section{Augmentation techniques} \label{Augmentation}

In this section we describe the data augmentation techniques we used.
We applied the same steps for each language with varying effect.
We performed ablation studies (Section~\ref{sec:ablation}) on some languages to investigate the individual effect of these techniques.

\subsection{Stem modification}

Neural networks tend to have difficulties with low amounts of training data as is the case with low resource languages.
For example a model trained on a language with 50-150 examples will learn to output the training character sequences.
In order to avoid this we used the data hallucination technique introduced in \citet{anastasopoulos-neubig-2019-pushing}, who identified the ``stem'' based on common substrings in the inflected forms of the same lemma.
Then they replace some characters of the stem with random characters.
We use a similar method but instead of using random characters, we sample them according to the unigram distribution of each language.
This way we created 10000 additional examples for each language in the training set.

\subsection{Copy}

Another attempt was to help the model learn to copy because without it the model can output wrong characters for the stem instead of copying it.
We added a maximum number of 10000 examples to the training set where the additional data for each language looks like: \\
\texttt{\small{izar izar    Lang-family;Lang-code;COPY}} \\
Copy is a new tag we added just for this specific task.

\begin{figure}[h]
\centerline{\includegraphics[scale=.38]{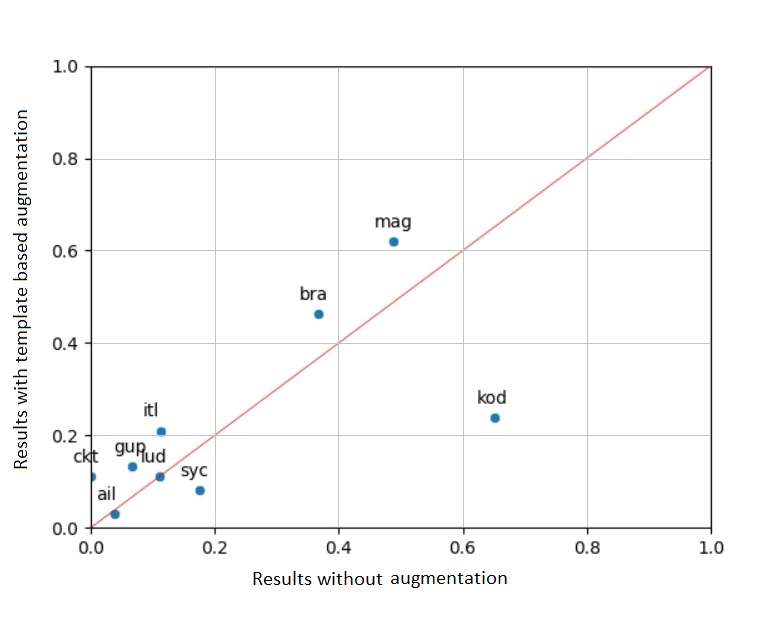}}
\caption{The results of the augmentation based on the work of \citet{Neuvel:2002}, evaluated on underresourced languages.
The majority of these results show that this kind of augmentation is not helpful.}
\label{fig}
\end{figure}

\subsection{Template based augmentation}

We also experimented with a morpheme identifier algorithm of \citet{Neuvel:2002}.
The main idea of \citet{Neuvel:2002} is to iteratively find ``similar'' substrings between inflected forms within the same languages.
Common substrings are considered affixes and they can be used in templates to generate further examples.
We first identify whether the language is more likely to use prefixes or suffixes.
We then apply the forward-comparing or the backward-comparing method to extract the affixes.
The forward-comparing method is explained in Algorithm~\ref{alg:FCM}.
The backward-comparing method is identical except it starts at the end of the string.


\begin{algorithm}
\caption{Forward-comparing method}\label{alg:FCM}
\begin{algorithmic}
\Require initialize 3 empty lists: sim, diff1, diff2
\For{$i\gets 0,  \min{\{len(w1), len(w2)}$\}}
    \If{w1[i] equals w2[i]}
        \State $sim\gets w1[i]$
        \State $diff1\gets "."$
        \State $diff2\gets "."$
    \Else
        \State $sim\gets "."$
        \State $diff1\gets w1[i]$
        \State $diff2\gets w1[i]$
    \EndIf
\EndFor
\State $sim\gets \|len(w1)-len(w2)\|$ "." characters
\State{\textbf{Append} the rest of the remaining words to the appropriate diff list}
\State \textbf{return} $diff1, diff2, sim$
\end{algorithmic}
\end{algorithm}

We store these 3 lists in a single data structure called ``comparison''.
We collect comparisons from all the records of our source data.
Afterwards we try to merge them based on their similarities.
Two comparisons are considered similar if in all of their three lists they only differ in '.' characters. If we find such comparisons, the differing '.' characters in all lists are replaced by '?' marks - creating a new comparison and the source comparisons are deleted. This merged comparison then can be considered for further merging too.
We also count how many merges happen in a comparison.
The results of this algorithm can be used to generate newer words by using diff1 and diff2 of a comparison as templates.
Starting from the left side of the template we generate exactly one character to replace each '.' characters (this one generated character is used to replace the holders in both diff1 and diff2), and one or zero character to replace each '?' marks.
We tried improving our results by generating the characters based on the frequency of the characters of a given source.
We also tried improving under resourced language results by using words generated by templates from the languages' families. 

The results with and without the template based augmentation are compared in Figure~\ref{fig}.
We only include a handful of low resource languages for clarity.
Template based augmentation is useful for some languages but it decreased the performance for others and it was computationally expensive to try models with and without template based augmentation for all languages.

\section{Experiments}

We first experimented with a single large model for all languages.
This model worked reasonably well for large languages and some of the medium ones as well, however, its performance was quite low on low resourced languages.
To offset this, we introduce a three-step training method.

\paragraph{Step 1: General training.}
In the first phase, we train a model on all languages with \emph{copy} augmentation.
This step creates a general language model.
Using \emph{copy} alone considerably improves the results on most languages.
We further explore the effect of \emph{copy} in Section~\ref{sec:ablation}.
We trained a small and a large model as described in Section~\ref{sec:model}.

\paragraph{Step 2: Language family training.} We fine-tuned both model for each language family but this time we also added 10000 example to each language with using the \emph{stem modification} augmentation.
This resulted in 24 models, one small and one large for each of the 12 language families.

\paragraph{Step 3: Language specific training.}
We then further fine-tune the family models on each language of the particular family.
We use template based augmentation in this step to further increase the number of training samples.

\section{Results}

Our three step model achieves 81.51\% accuracy on average with 0.627 average Levenshtein distance between the output and the ground truth.
This results is lower than the baseline (84.57\% accuracy and 0.448 edit distance) but it is important to note that our LSTM seq2seq model is simpler than the Transformer baseline, while it remains competitive.
We chose our model based on the accuracy of the development dataset. This way we submitted the results of 7 smaller and 31 larger models.

\subsection{Ablation studies}
\label{sec:ablation}

We perform a small ablation studies on two augmentation technique and on each step of training process.
We picked all languages where our results were better than the baseline and retrained the model without one of the augmentation techniques or without one of the three training steps.
The results are listed in Table~\ref{tab:results}.
It is clear that \emph{copy} is the most important data augmentation technique.
\emph{Stem modification} often has a mixed effect on these languages.
The model on Ludic, the smallest language in the data set, is actually much better without stem modification.
The same is true about the training steps.
The overall accuracy increased with each step but the language-specific effects are more varied.
Since testing with and without all options is computationally too expensive for such a large number of languages, we recommend trying these options individually for language-specific applications.



\section{Conclusion}

We presented the BME team's submission for the SIGMORPHON 2021 Task 0 Part 1 shared task.
We used a standard LSTM encoder-decoder model with attention.
We augmented the data with three different techniques and then used a three-step training method that first trains on all languages, then fine-tunes on the language families and finally fine-tunes on the individual languages.
Our submission outperformed the other team by a large margin.
Although our results did not outperform the baseline, we presented a simpler model with multiple data augmentation techniques that do not require any external resource or linguistic expertise.

\bibliographystyle{acl_natbib}
\bibliography{acl2021}


\end{document}